\documentclass{article} % For LaTeX2e
\usepackage{iclr2024_conference,times}

\usepackage{amsmath,amsfonts,bm}

\def\eqref#1{equation~\ref{#1}}
\def\1{\bm{1}}

\DeclareMathAlphabet{\mathsfit}{\encodingdefault}{\sfdefault}{m}{sl}
\SetMathAlphabet{\mathsfit}{bold}{\encodingdefault}{\sfdefault}{bx}{n}

\usepackage{hyperref}
\usepackage{url}
\usepackage{graphicx}
\usepackage{caption}
\usepackage{subcaption}
\usepackage{multirow}
\usepackage{array}
\usepackage{float}

\title{Fine-Tuning Small Embeddings for Elevated Performance}

\author{Biraj Silwal\\
Department of Electronics and Computer Engineering\\
Institute of Engineering, Pulchowk Campus\\
Lalitpur, Nepal \\
\texttt{078msdsa005.biraj@pcampus.edu.np} \\
}

\iclrfinalcopy % Uncomment for camera-ready version, but NOT for submission.
\begin{document}

\maketitle

\begin{abstract}
Contextual Embeddings have yielded state-of-the-art results in various natural language processing tasks. However, these embeddings are constrained by models requiring large amounts of data and huge computing power. This is an issue for low-resource languages like Nepali as the amount of data available over the internet is not always sufficient for the models. This work has taken an incomplete BERT model with six attention heads pretrained on Nepali language and finetuned it on previously unseen data. The obtained results from intrinsic and extrinsic evaluations have been compared to the results drawn from the original model baseline and a complete BERT model pretrained on Nepali language as the oracle. The results demonstrate that even though the oracle is better on average, finetuning the small embeddings drastically improves results compared to the original baseline. 
\end{abstract}

\section{Introduction}
The recent advancement in computation technologies have yielded state-of-the-art performances in multiple Natural Language Processing(NLP) tasks such as Text Classification, Named-Entity Recognition, Question Answering and Sentiment Analysis. The techniques used to resolve these problems revolve around word representation and the concept of representing human understanding of the language in a machine recognizable form. The default approach of representing words as discrete and distinct symbols is insufficient for many tasks, and suffers from poor generalization.\citep{cite1} Thus, representation of words plays a vital role in the resolution of NLP problems.

Word Embedding is one of the most widely used technique for word representation. In natural language processing (NLP), word embedding is a term used for the representation of words for text analysis, typically in the form of a real-valued vector that encodes the meaning of the word such that the words that are closer in the vector space are expected to be similar in meaning.\citep{cite2} Word Embeddings can generally be divided into two categories: Context-Independent embeddings such as GloVe \citep{pennington2014glove}, Word2Vec \citep{mikolov2013distributed}, and fastText \citep{bojanowski2017enriching}{}, and Context-Dependent embeddings such as BERT (Bidirectional Encoder Representations from Transformers) \citep{devlin2018bert} and ELMo (Embeddings from Language Models) \citep{peters2018deep}. Context-Independent word embeddings are generated for words as a function whereas Context-Dependent word embeddings are generated for words as a function of the sentence it occurs in. 

Context-Independent word embeddings have the major drawback of conflating words with various meanings into a single representation. Context-Independent word embeddings, on the other hand, better capture the multi-sense nature of words as they are at the token level and each occurrence of a word has its own embedding. To capture these contextualized representations, BERT uses a transformer that has been trained on tasks like Next Sentence Prediction and Masked Language Modeling. BERT has become really popular in NLP tasks even though initially there were only two pre-trained BERT versions—one in Chinese and one in English.

The major issue with context-dependent embedding is the amount of computing resources and data required to produce effective results. Although Nepali is a language spoken by millions of people, it is a complex language with rich vocabulary and diverse grammatical structures with low resource availability on the internet. When these two issues are put together, we have the problem of a data hungry model receiving insufficient data for processing. To address this issue, an incomplete BERT model has been taken and fine-tuned on previously unseen unregularized corpus of Nepali text in order to generate word embeddings that better capture the semantic and syntactic relationships between words in Nepali sentences.

The generated word embeddings can be used for a variety of NLP tasks, such as text classification, named entity recognition, and sentiment analysis. They can also be used to improve machine translation systems for Nepali language, which would have a significant impact on communication and information sharing in the region. Overall, the goal of this work was to amplify the power of pre-trained smaller language models to improve the quality of word embeddings for Nepali language and enable the development of more accurate and effective NLP applications.

\section{Related Works}
With the introduction of multilingual BERT, multiple researches have been done on NLP tasks in Nepali language using BERT. NepBERTa: Nepali Language Model Trained in a Large Corpus\cite{cite8} presented a BERT-based Natural Language Understanding (NLU) model trained on the most extensive monolingual Nepali corpus ever. NepBERTa's performance was assessed in several Nepali-specific NLP tasks, including Named-Entity Recognition, Content Classification, POS Tagging, and Categorical Pair Similarity. Additionally, two new datasets were introduced for  two new downstream tasks and these four tasks were brought together as the first-ever Nepali Language Understanding Evaluation (Nep-gLUE) benchmark.

NPVec1: Word Embeddings for Nepali - Construction and Evaluation \citep{cite3} introduced twenty five state-of-art Word Embeddings for Nepali derived from a large corpus using GloVe, Word2Vec, fastText, and BERT. However, the BERT model was only trained in one preprocessing scheme and made up only one of the twenty five word embeddings. Also, the BERT architectucture used was trained only on 360 million words while the original model was trained on 3.3 billion words.

NepaliBERT\citep{nepaliBERT} was developed using a training set of 85467 news scrapped from different job portals. The corpus size was about 4.3 GB of textual data. Similarly, the evaluation data contained news articles of about 12 MB of textual data. At the time of training, this state of the art model demonstrated an Intrinsic evaluation with Perplexity of 8.56 and extrinsic evaluation performed on a downstream task i.e sentiment analysis of Nepali tweets outperformed other existing masked language models.

%% Please note that we have introduced automatic line number generation
%% into the style file for \LaTeXe. This is to help reviewers
%% refer to specific lines of the paper when they make their comments. Please do
%% NOT refer to these line numbers in your paper as they will be removed from the
%% style file for the final version of accepted papers.

\section{Methodology}
\subsection{Data Collection}
As was previously mentioned, a significant amount of data would be needed to generate word embeddings using BERT. Data is the project's foundation, so to speak. Despite being scarcely present online, Nepali data can be found in a variety of formats. These data can be found in a variety of places, including Nepali news articles published by online news sources, Nepalese websites, Nepal Government websites, social media posts in Nepali, and more. The data for this study has been broadly divided into two sections: Regularized data and Unregularized data. 

\subsubsection{Collection of Regularized Data}
The category of Regularized Data generally encompasses data which may come from sources with a greater degree of editing and reviews. Online news articles are the primary example of regularized data. This is due to the fact that, to publish an article on the website, the article must go through multiple drafts while being reviewed and edited by a certified editorial board. Due to these reviews, the language used in these platforms follow a general rule and may lose the nuances of the Nepali language. 

Nepali news portals like Onlinekhabar, Ratopati, etc are major sources of Nepali language data on the internet. The continuous updates on those portals and availability of articles for a substantial period of time, has accounted for a major data source for the project. The online news portals were first selected and inspected, to verify the viability of scraping. Then, a web scraping script was written in Python with the help of the BeautifulSoup library to extract the required information from the web page. The scraped text was then stored in a .txt file. The process was further automated using the Selenium library and repeated for a large number of web pages to scrape the required data.

\subsubsection{Collection of Unegularized Data}
Along with Regularized Data, the internet is also a great source of Unregularized Data. In a general sense, Unregularized Data can be understood as data which was deemed to originate from sources with a lesser degree of editorial review. Social media posts are the best example to illustrate Unregularized Data. The influx of users on sites like Facebook and Twitter has produced a new source of text and visual data. These data generally come directly from a personal standpoint without interference from layers of editors. Thus, this data is better able to grasp the nuances of the Nepali language. 

Primarily, various social media sites were observed to find the best fit for data collection. Sites like Facebook, Twitter and Reddit had occurences of Nepali language, either in posts made by some users, comments posted and even in identifiers of individual accounts i.e usernames. Facebook and Twitter were selected as the best fits due to the abundance of Nepali text data and public access to their respective Application Programming Interfaces. The data was extracted from Facebook using the Facebook Graph API from a Facebook Developer account and from Twitter using the Twitter API and the tweepy library in python. The data collected was then stored in a database for further use.

\subsection{Preprocessing of Aggregated Corpus}
The collected data were then evaluated to prepare them for further processing. The Regularized and Unregularized data were aggregated and various techniques were used during the preprocessing stage.

\subsubsection{Filtering Non-Nepali Data}
The major problem with the aggregated corpus was the occurence of Hindi data instances within the Nepali data. This issue observed, was found to be a direct consequence of the use of Facebook Graph API and the twitter API to extract data. Due to both languages being written in Devanagari script, the search results for tweets or posts in Nepali language would also incorporate Hindi data. To resolve this issue, a python library named langdetect was used. 

Langdetect was able to detect 55 languages and Hindi was one of them. The main idea behind this was to detect the Hindi data and remove them from the corpus. In contrast to the expectations, it was observed that all data instances were marked as Hindi. To overcome this issue, removal was done with the use off regular expressions. A pattern of 100 most occurring Hindi words was created and the data instances were matched against the pattern. Any match found was filtered out to generate the aggregated Nepali corpus.

\subsubsection{Standardization}
 In Nepali, there are various written vowel sounds that sound identical when spoken, similar to how there are different cases (lower/upper) in English without any phonetic variations. To avoid multiple occurrences of the same token in the corpus, the aggregated corpus was standardized to remove the instances of multiple Nepali vowels. The process of removing these multiple instances was done by creating an index of Nepali vowel sounds and reducing a pair of those vowels to a singular vowel. This preprocessing technique has been used to help  reduce noise in the data by eliminating any instances of miswritten Nepali data.
 
\subsubsection{Lexical Analysis}
Nepali is an agglutinative language i.e there are numerous post-positional suffixes in the Nepali language that can be combined with nouns and pronouns to form new terms. The Lexical analysis is performed by using a tokenizer which breaks the words down into tokens of the base word and the post-positional suffixes. This leads to multiple instances of different words being broken down into overlapping set of tokens. Lexical analysis, thus helps to reduce the vocabulary by breaking the words into tokens of the base word and the respective post-positional suffixes.

\begin{figure}[H]
\centering
\includegraphics[scale = 0.5]{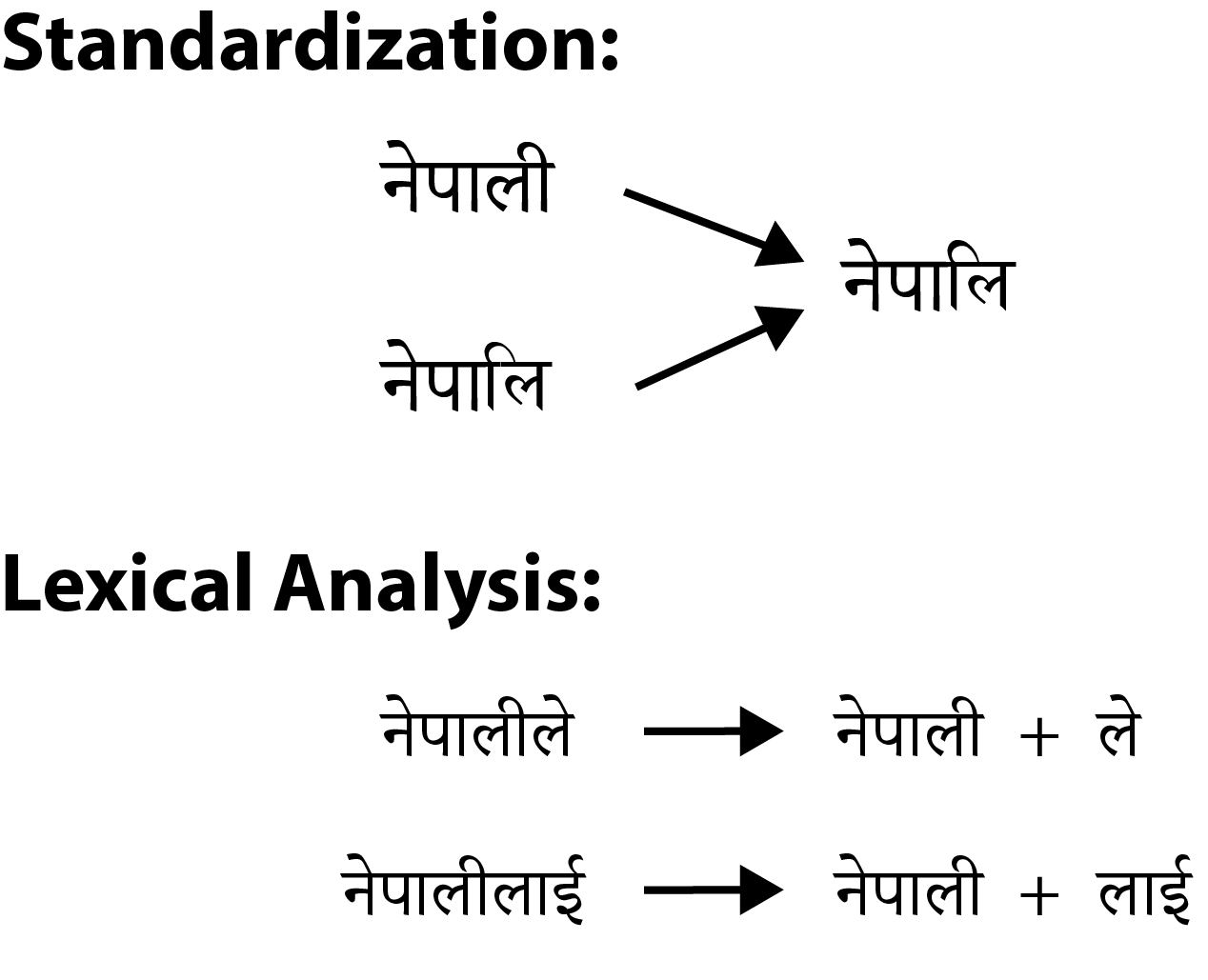}
\caption{\label{fig:flowchart} Preprocessing techniques}
\end{figure}

\subsection{Transfer Learning for Word Embedding Generation on BERT based Models}
Transfer learning is a machine learning technique where a model trained on one task is fine-tuned on a second related task. This allows the model to benefit from the knowledge it has gained while solving the first task and apply it to the second task. The idea is to leverage the features learned by the model on the first task as a starting point, rather than starting from scratch. Transfer learning with BERT involves fine-tuning the pre-trained BERT model on a specific NLP task by adding task-specific layers on top of the BERT layers. The fine-tuning process adjusts the weights of the BERT model to better fit the new task. Unlike training a model from scratch, transfer learning with BERT allows the model to take advantage of the knowledge learned from the pre-training process and fine-tune it to the specific NLP task more efficiently.

 In this project, the model generated from NPVec1: Word Embeddings for Nepali has been used as the base model. This model has been trained using 279 million word tokens and is the largest Embeddings ever trained for Nepali language. This model consists of 6 attention heads and 300 hidden dimensions. This project has taken the NPVec1 BERT model as the base and fine-tuned the model to accommodate the generated smaller corpus. This enabled the introduction of an unregulated dataset to a BERT model trained primarily on data from news websites. The model has further been trained on various iterations of preprocessing techniques applied to the raw data. Finally, the generated word embeddings have been used for intrinsic and extrinsic evaluation.

\section{Evaluation}
\subsection{Corpus Description}
The corpus was generated after a few ruls of the scraping script for both regularized and unregularized data followed by the preprocessing steps. The description of the Regulated and Unregulated corpora and the aggregated corpus is given in figure \ref{corpus description}.

\begin{table}[H]
    \centering
    \begin{tabular}{c c c}
    \hline
         Corpus Type & Word Token Count & Word Type Count  \\
         \hline
         Regualted Corpus & 43.58M & 0.38M \\
         Unregulated Corpus & 96.90M & 1.25M\\
         Aggregated Corpus & 140.48M & 1.40M\\
         \hline
    \end{tabular}
    \caption{Corpus Description}
    \label{corpus description}
\end{table}

\subsection{Intrinsic Evaluation}
As said previously, the Intrinsic Evaluation of the embeddings was done by the method of clustering. To accommodate the metrics for comparison, the evaluation was done for the finetuned NpVec1 model, NpVec1 model and the nepaliBERT huggingface transformers. The NpVec1 model has been assumed as the baseline model as it is the precursor to the finetuned model and the nepaliBERT model has been assumed to be the oracle as the model is based on the original BERT implementation of 12 hidden layers and attention heads and 768 hidden dimensions. 

\begin{table}[H]
    \centering
    \begin{tabular}{c c c c c}
        \hline
         \multirow{2}{3em}{Model} & \multicolumn{4}{c}{Purity}  \\
         \cline{2-5}
          & Sentiment & Relatedness & Named Entity & Average\\
          \hline
          NpVec1 & 0.53 & 0.82 & 0.60 & 0.65 \\
          Finetuned Model & 0.76 & 0.77 & 0.80 & 0.78\\
          nepaliBERT & 0.73 & 0.80 & 0.92 & \textbf{0.82}\\
         \hline
    \end{tabular}
    \caption{Results of Intrinsic Evaluation. Scores represent purity values on a scale of 0 to 1.}
    \label{intrinsic evaluation}
\end{table}

The results of the intrinsic evaluation has been presented in table \ref{intrinsic evaluation}. The results clearly demonstrates the positive effect finetuning has on the embeddings. It was seen that although the Finetuned model was outperformed by the other models in the Relatedness and Named Entity sets, the deviation from the highest score was only 5\% and 12\% respectively. Also, the constant performance of the Finetuned model across all three sets indicated that it had a better completeness across domains. Furthermore, the deviation with its precursor i.e NpVec1 model was seen to be a 23\% lead, 5\% lag and a 20\% lead in the Sentiment set, Relatedness set and Named Entity set respectively. On average, the finetuned model performs better than the precursor NPVec1 model by a considerable margin. 

\subsection{Sentiment Clusters}
\begin{figure}[H]
\begin{minipage}{0.33\linewidth}
\includegraphics[width=\textwidth]{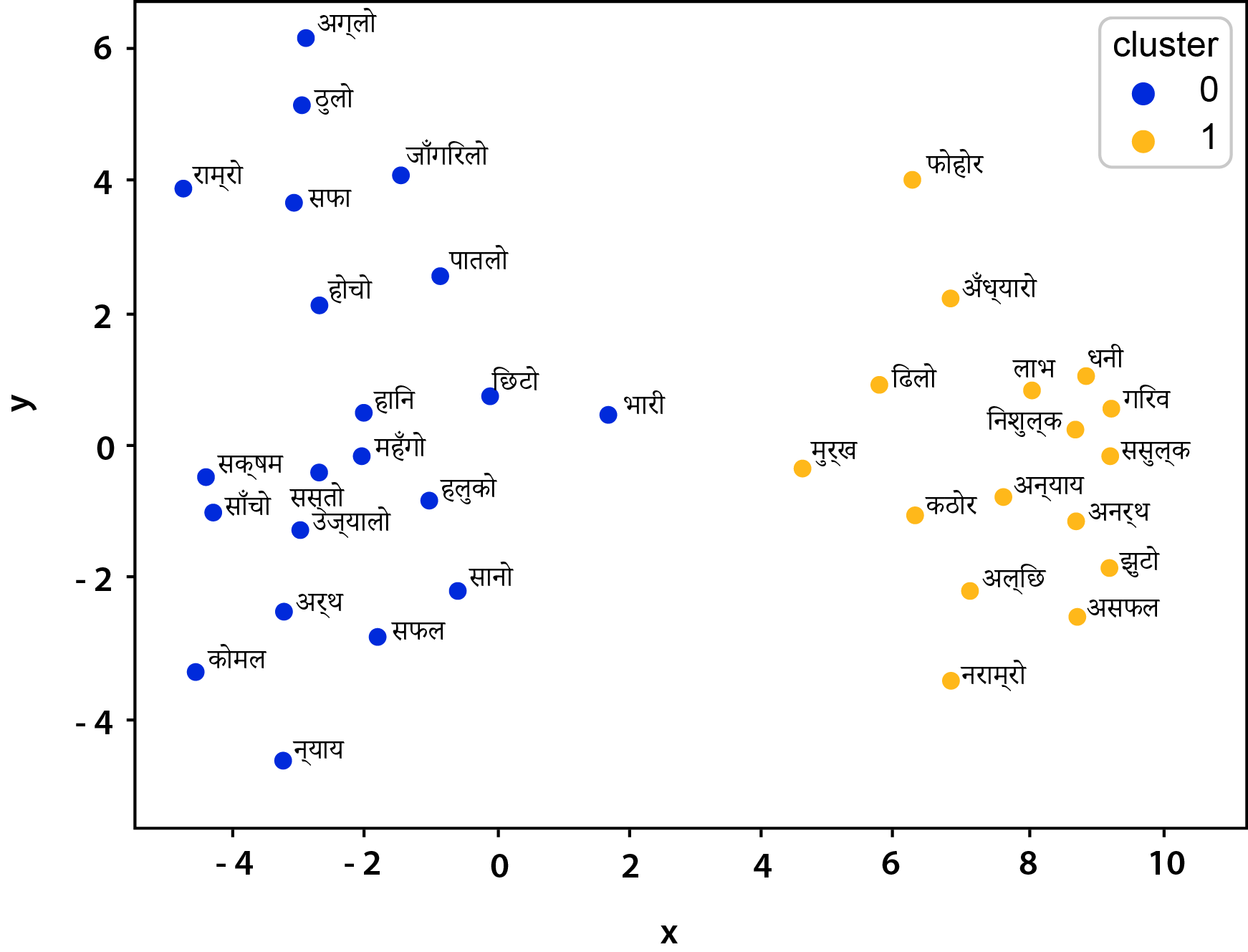}
\caption{Finetuned model}
\label{Finetuned model}
\end{minipage}%
\hfill
\begin{minipage}{0.33\linewidth}
\includegraphics[width=\textwidth]{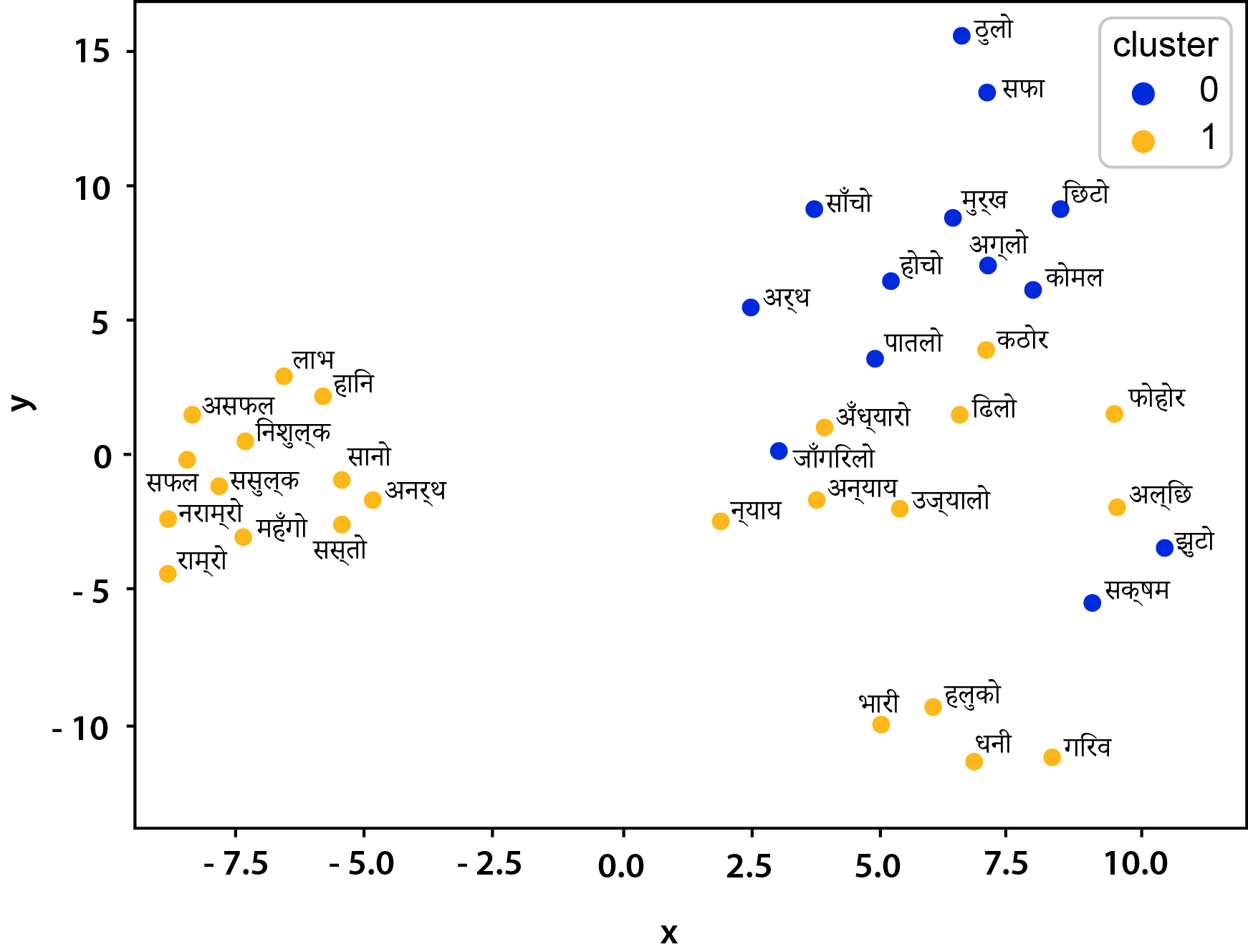}
\caption{NpVec1}
\label{NpVec1}
\end{minipage}%
\hfill
\begin{minipage}{0.33\linewidth}
\includegraphics[width=\textwidth]{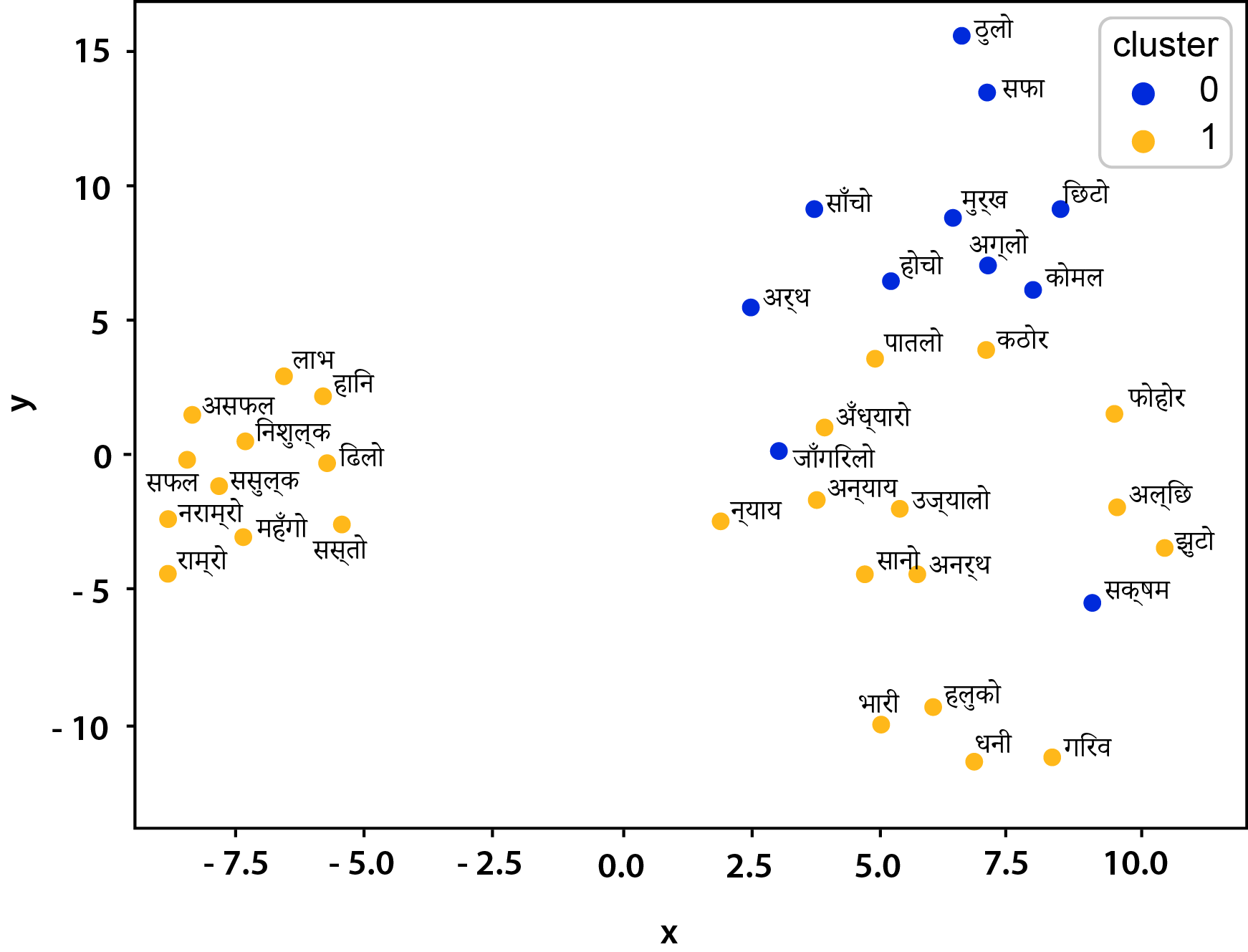}
\caption{nepaliBERT}
\label{nepaliBERT}
\end{minipage}
\end{figure}

\subsection{Relatedness Clusters}
\begin{figure}[H]
\begin{minipage}{0.33\linewidth}
\includegraphics[width=\textwidth]{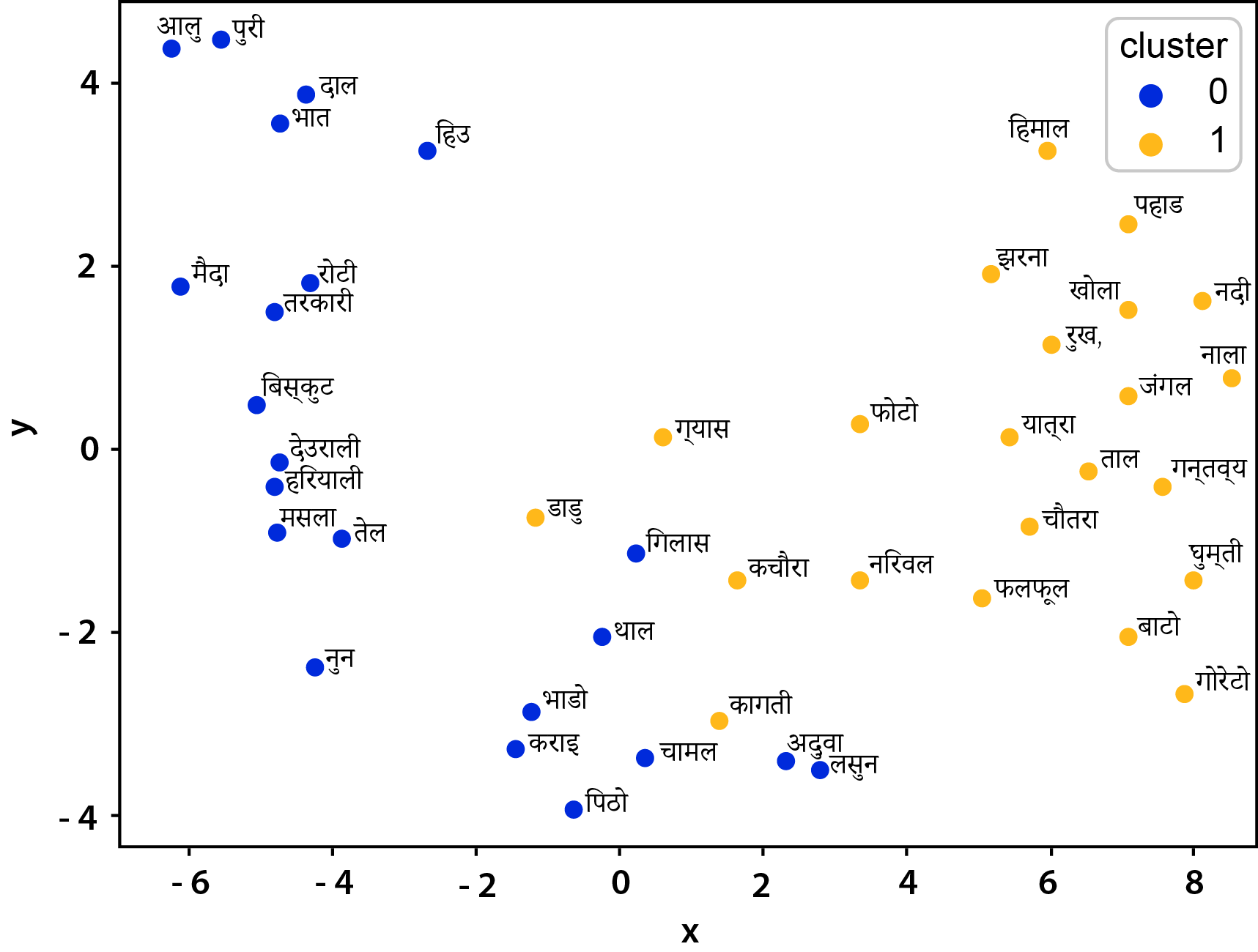}
\caption{Finetuned model}
\label{fig:figure1}
\end{minipage}%
\hfill
\begin{minipage}{0.33\linewidth}
\includegraphics[width=\textwidth]{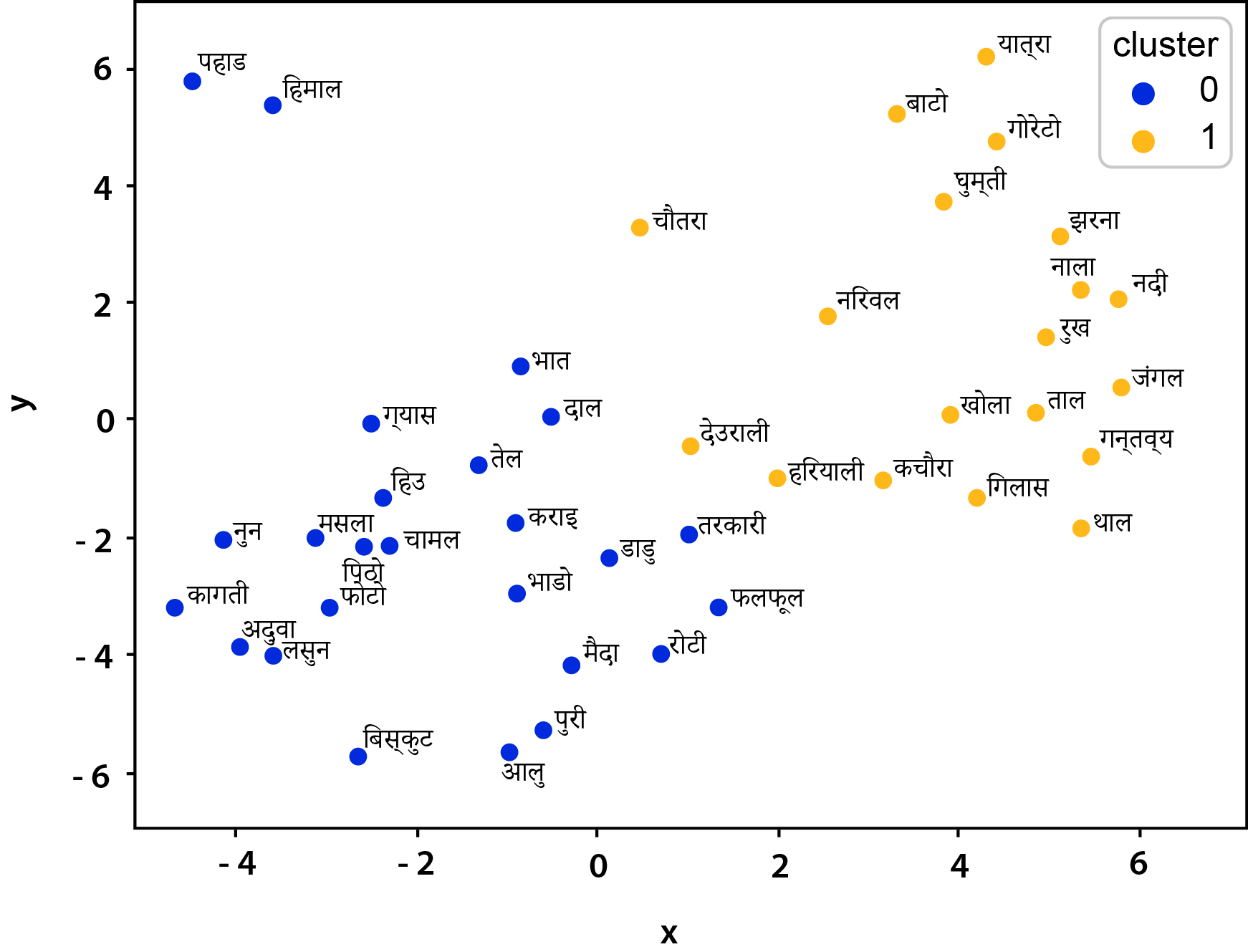}
\caption{NpVec1}
\label{fig:figure2}
\end{minipage}%
\hfill
\begin{minipage}{0.33\linewidth}
\includegraphics[width=\textwidth]{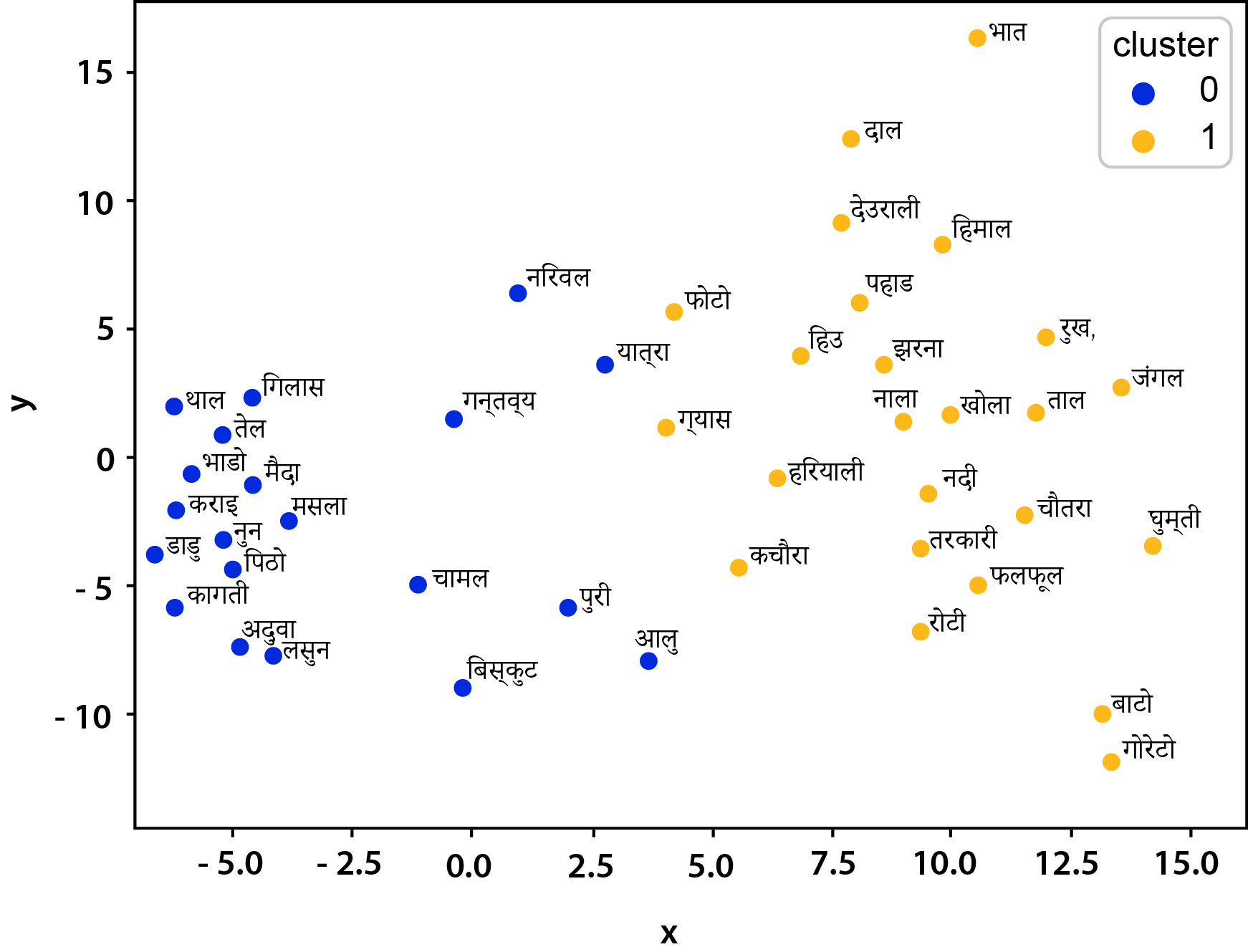}
\caption{nepaliBERT}
\label{fig:figure3}
\end{minipage}
\end{figure}

\subsection{Named Entity Clusters}
\begin{figure}[H]
\begin{minipage}{0.33\linewidth}
\includegraphics[width=\textwidth]{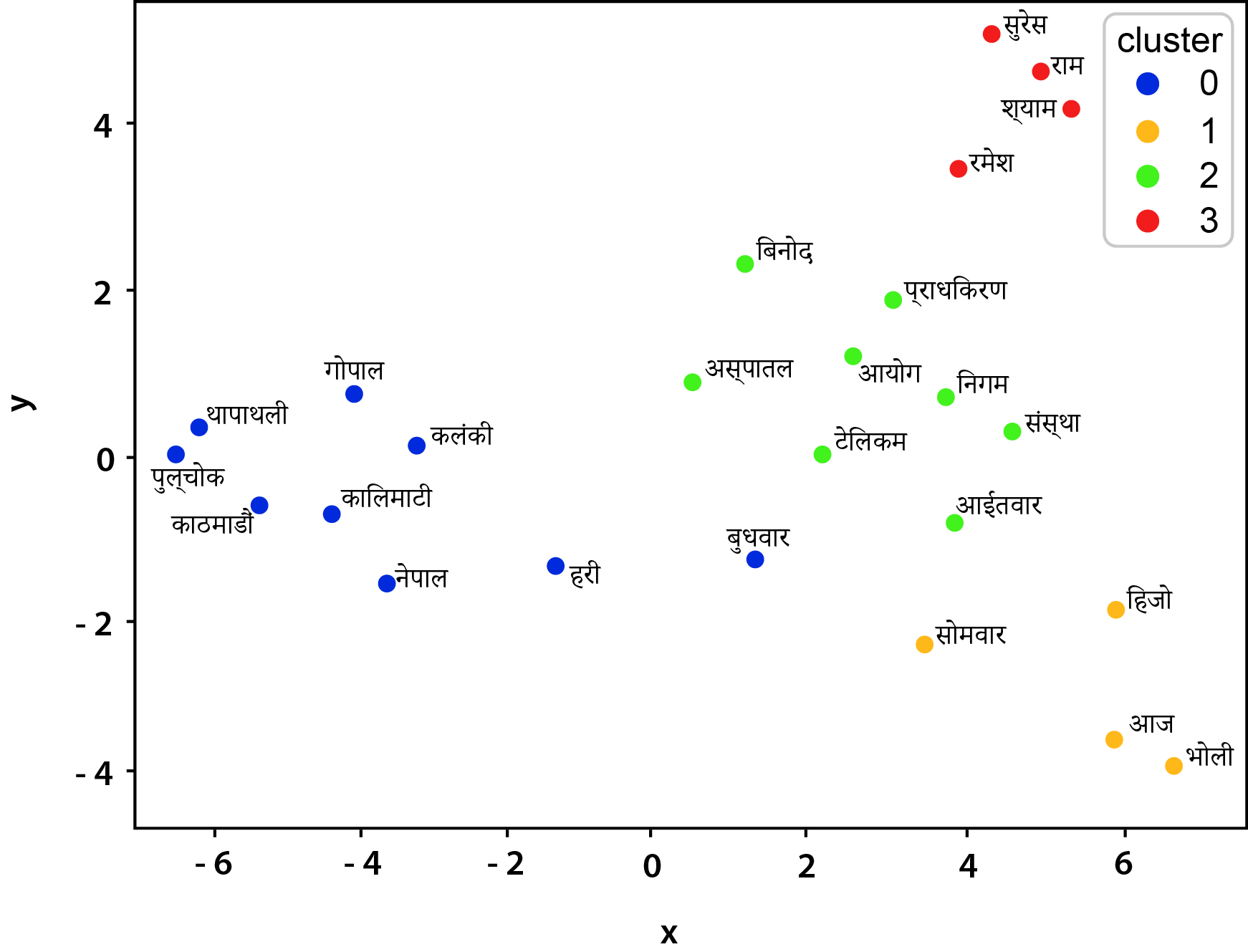}
\caption{Finetuned model}
\label{fig:figure1}
\end{minipage}%
\hfill
\begin{minipage}{0.33\linewidth}
\includegraphics[width=\textwidth]{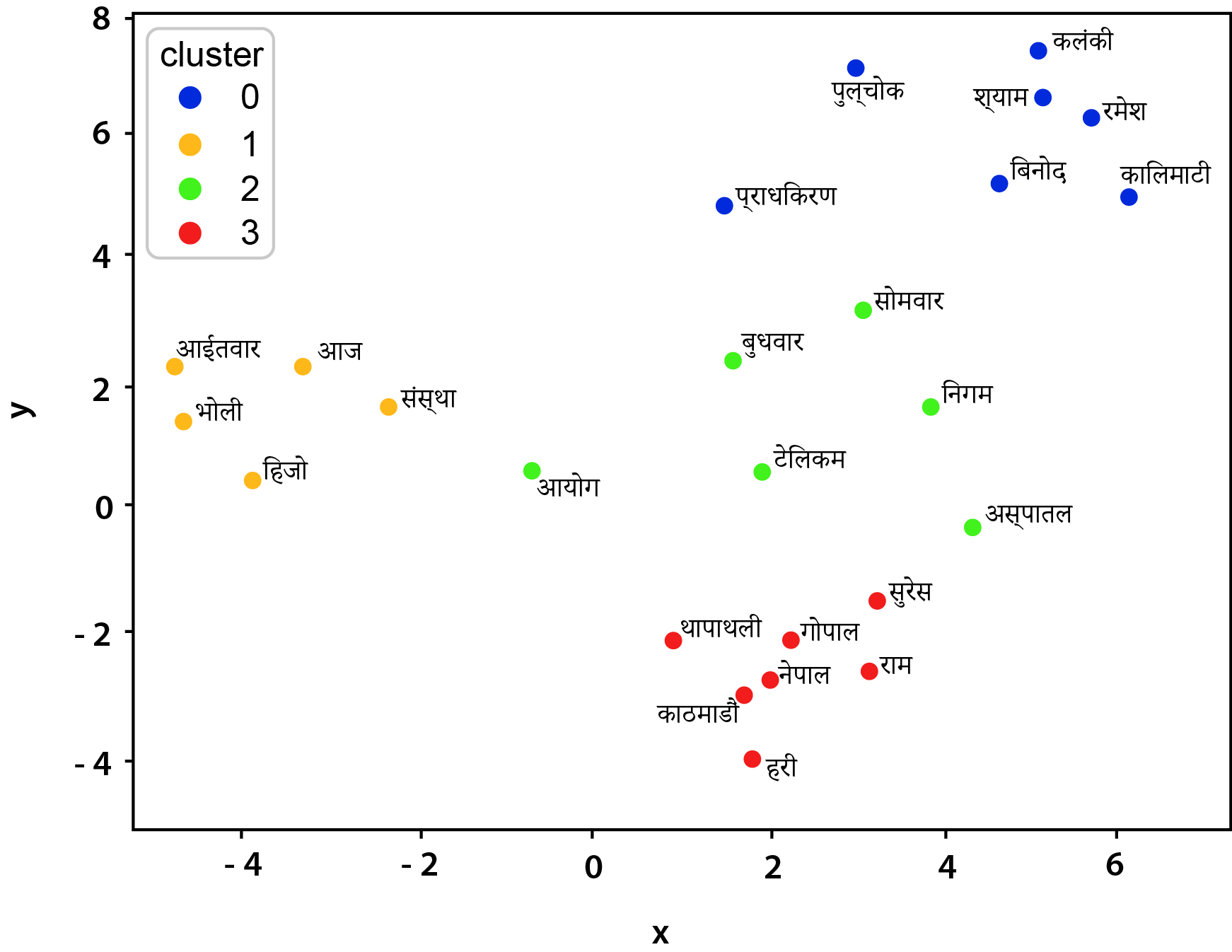}
\caption{NpVec1}
\label{fig:figure2}
\end{minipage}%
\hfill
\begin{minipage}{0.33\linewidth}
\includegraphics[width=\textwidth]{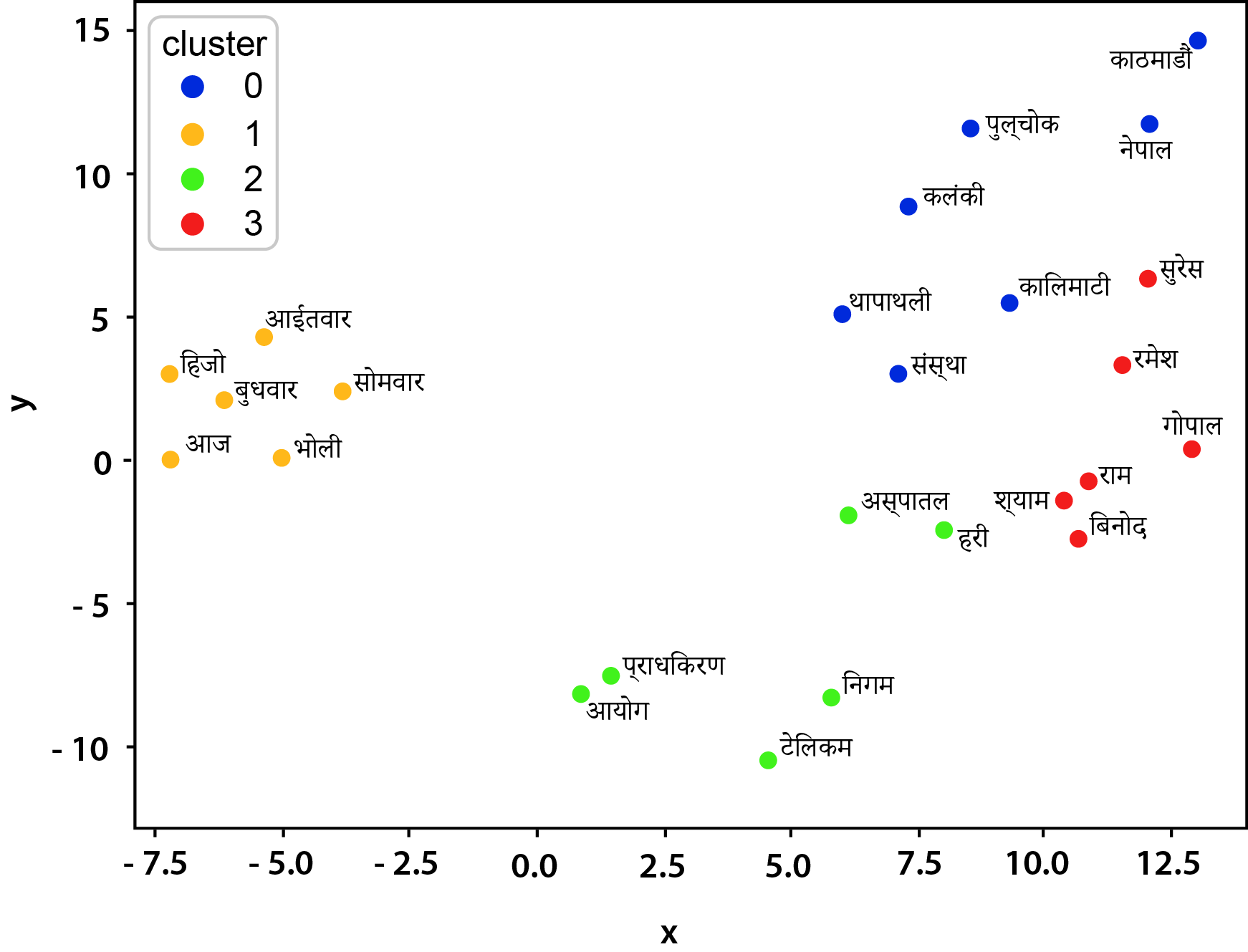}
\caption{nepaliBERT}
\label{fig:figure3}
\end{minipage}
\end{figure}

\subsection{Extrinsic Evaluation}
The Extrinsic Evaluation was performed on a news classification task as stated previously. The evaluation process was done in 60 epochs and it was done for the finetuned model, nepaliBERT model and the NpVec1 model respectively. The reason for using the three models was to evaluate the performance of the finetuned model generated by this study with its precursor NpVec1 as the baseline and the bigger model nepaliBERT as the upper bound. The following results were obtained for the extrinsic evaluation.

\begin{table}[h]
    \centering
    \begin{tabular}{c c c c}
        \hline
         Model & Precision & Recall & F1-Score \\
         \hline
         NpVec1 & 0.73 & 0.74 & 0.74 \\
         Finetuned Model & 0.82 & 0.79 & 0.81 \\
         nepaliBERT & \textbf{0.86} & \textbf{0.88} & \textbf{0.87}\\
         \hline
    \end{tabular}
    \caption{Performance of models in Extrinsic Evaluation task}
    \label{tab:my_label}
\end{table}

The classification model was assessed using macro precision, recall, and F1 measures. It was clearly observed that the model built on the original BERT model i.e NepaliBERT easily outperforms the the other models. Even though the models NepaliBERT and NpVec1 have been trained on similar volumes of data, the difference in performance of 13\%, with respect to F1-score, is substantial.

For the main purpose of this study, we can see that the performance of the Finetuned Model is significantly better than that of its precursor i.e NpVec1. Although NpVec1 was pre-trained on approximately 4 times more data, the fine-tuning performed using a smaller dataset helped improve the models performance by 7\%, with respect to the F1-score. The finetuned model also outperfomed the NpVec1 model on both Precision and Recall metrics demonstrating that the finetuned models results were more accurate and more complete than the results of the NpVec1 model respectively.

\section{Conclusion}
 The main objective of this study was to generate the word embeddings by fine-tuning a existing BERT model and evaluate them using various metrics. The metrics were then compared against some existing pre-trained BERT models. With the above evaluations, it was observed that although the Finetuned model was built on a significantly lower amount of data on top of the NpVec1 model, it substantially outperformed the model in both Intrinsic and Extrinsic evaluations. This might majorly be attributed to the use of Unregulated data which was missing during the pre-training of the NpVec1 model. Thus, it can be concluded that fine-tuning a pre-trained BERT model using a completely new unlabeled dataset can lead to substantially better performances than the pre-trained model; even sometimes competing with a model with greater architecture, in some of the metrics. This illustrates that, pre-training and finetuning a low-resource language model on a wide data domain can become a possible solution to being unable to pre-train low-resource language models due to lack of proper data.

\bibliography{iclr2024_conference}

\begin{thebibliography}{10}
\providecommand{\natexlab}[1]{#1}
\providecommand{\url}[1]{\texttt{#1}}
\expandafter\ifx\csname urlstyle\endcsname\relax
  \providecommand{\doi}[1]{doi: #1}\else
  \providecommand{\doi}{doi: \begingroup \urlstyle{rm}\Url}\fi

\bibitem[Bojanowski et~al.(2017)Bojanowski, Grave, Joulin, and Mikolov]{bojanowski2017enriching}
Piotr Bojanowski, Edouard Grave, Armand Joulin, and Tomas Mikolov.
\newblock Enriching word vectors with subword information.
\newblock \emph{Transactions of the association for computational linguistics}, 5:\penalty0 135--146, 2017.

\bibitem[Devlin et~al.(2018)Devlin, Chang, Lee, and Toutanova]{devlin2018bert}
Jacob Devlin, Ming-Wei Chang, Kenton Lee, and Kristina Toutanova.
\newblock Bert: Pre-training of deep bidirectional transformers for language understanding.
\newblock \emph{arXiv preprint arXiv:1810.04805}, 2018.

\bibitem[Jurafsky(2000)]{cite2}
Dan Jurafsky.
\newblock \emph{Speech \& language processing}.
\newblock Pearson Education India, 2000.

\bibitem[Koirala \& Niraula(2021)Koirala and Niraula]{cite3}
Pravesh Koirala and Nobal~B Niraula.
\newblock Npvec1: Word embeddings for nepali-construction and evaluation.
\newblock In \emph{Proceedings of the 6th Workshop on Representation Learning for NLP (RepL4NLP-2021)}, pp.\  174--184, 2021.

\bibitem[Levy \& Goldberg(2014)Levy and Goldberg]{cite1}
Omer Levy and Yoav Goldberg.
\newblock \emph{Dependency-based word embeddings}.
\newblock 2014.

\bibitem[Mikolov et~al.(2013)Mikolov, Sutskever, Chen, Corrado, and Dean]{mikolov2013distributed}
Tomas Mikolov, Ilya Sutskever, Kai Chen, Greg~S Corrado, and Jeff Dean.
\newblock Distributed representations of words and phrases and their compositionality.
\newblock \emph{Advances in neural information processing systems}, 26, 2013.

\bibitem[Pennington et~al.(2014)Pennington, Socher, and Manning]{pennington2014glove}
Jeffrey Pennington, Richard Socher, and Christopher~D Manning.
\newblock Glove: Global vectors for word representation.
\newblock In \emph{Proceedings of the 2014 conference on empirical methods in natural language processing (EMNLP)}, pp.\  1532--1543, 2014.

\bibitem[Peters et~al.(2018)Peters, Neumann, Iyyer, Gardner, Clark, Lee, and Zettlemoyer]{peters2018deep}
Matthew~E. Peters, Mark Neumann, Mohit Iyyer, Matt Gardner, Christopher Clark, Kenton Lee, and Luke Zettlemoyer.
\newblock Deep contextualized word representations, 2018.

\bibitem[Pudasaini et~al.(2023)Pudasaini, Shakya, Tamang, Adhikari, Thapa, and Lamichhane]{nepaliBERT}
Shushanta Pudasaini, Subarna Shakya, Aakash Tamang, Sajjan Adhikari, Sunil Thapa, and Sagar Lamichhane.
\newblock Nepalibert: Pre-training of masked language model in nepali corpus.
\newblock In \emph{2023 7th International Conference on I-SMAC (IoT in Social, Mobile, Analytics and Cloud) (I-SMAC)}, pp.\  325--330, 2023.
\newblock \doi{10.1109/I-SMAC58438.2023.10290690}.

\bibitem[Timilsina et~al.(2022)Timilsina, Gautam, and Bhattarai]{cite8}
Sulav Timilsina, Milan Gautam, and Binod Bhattarai.
\newblock Nepberta: Nepali language model trained in a large corpus.
\newblock In \emph{Proceedings of the 2nd Conference of the Asia-Pacific Chapter of the Association for Computational Linguistics and the 12th International Joint Conference on Natural Language Processing}, pp.\  273--284, 2022.

\end{thebibliography}
\bibliographystyle{iclr2024_conference}
\end{document}